\documentclass[a4paper,twoside]{article}
\usepackage{graphics}
\usepackage{graphicx}
\usepackage{epsfig}
\usepackage{subcaption}
\usepackage{calc}
\usepackage{amssymb}
\usepackage{amstext}
\usepackage{amsmath}
\usepackage{amsthm}
\usepackage{multirow}
\usepackage{booktabs}
\usepackage{float}
\usepackage{multicol}
\usepackage{pslatex}
\usepackage{apalike}
\usepackage{cleveref}
\newcommand{\ra}[1]{\renewcommand{\arraystretch}{#1}}
\usepackage{SCITEPRESS}     % Please add other packages that you may need BEFORE the SCITEPRESS.sty package.

\begin{document}

\title{Early Bird: Loop Closures from Opposing Viewpoints for Perceptually-Aliased Indoor Environments}

% \author{\authorname{First Author Name\sup{1}\orcidAuthor{0000-0000-0000-0000}, Second Author Name\sup{1}\orcidAuthor{0000-0000-0000-0000} and Third Author Name\sup{2}\orcidAuthor{0000-0000-0000-0000}}
% \affiliation{\sup{1}Institute of Problem Solving, XYZ University, My Street, MyTown, MyCountry}
% \affiliation{\sup{2}Department of Computing, Main University, MySecondTown, MyCountry}
% \email{\{f\_author, s\_author\}@ips.xyz.edu, t\_author@dc.mu.edu}
% }

\author{\authorname{Satyajit Tourani\sup{*1}, Dhagash Desai\sup{*1}, Udit Singh Parihar\sup{*1}, Sourav Garg\sup{2}, Ravi Kiran Sarvadevabhatla\sup{3},  Michael Milford\sup{2} and K. Madhava Krishna\sup{1}}
\affiliation{\sup{*}Denotes authors with equal contribution}
\affiliation{\sup{1}Robotics Research Center, IIIT Hyderabad.}
\affiliation{\sup{2}Centre for Robotics, Queensland University of Technology (QUT), Australia.}
\affiliation{\sup{3}Centre for Visual Information Technology, IIIT Hyderabad.}
%\email{\{satyajit.tourani@research.iiit.ac.in\}}
%}
\email{\{satyajit.tourani@research.iiit.ac.in, dhagash.1@iitj.ac.in, uditsinghparihar96@gmail.com, s.garg@qut.edu.au, ravi.kiran@iiit.ac.in, michael.milford@qut.edu.au, mkrishna@iiit.ac.in\}}
 }

% \author{Satyajit Tourani$^{*1}$, Dhagash Desai$^{*1}$, Udit Singh Parihar$^{*1}$, Sourav Garg$^{2}$, \\ Ravi Kiran Sarvadevabhatla$^{3}$, Michael Milford$^{2}$ and K. Madhava Krishna$^{1}$%

% \affiliation{*Equal contribution}% <-this % stops a space
% \affiliation{$^{1}$Robotics Research Center, IIIT Hyderabad \\ $^{2}$Centre for Robotics, Queensland University of Technology (QUT), Australia. \\ $^{3}$Centre for Visual Information Technology, IIIT Hyderabad.}
% }

% \keywords{Visual Place Recognition, Vision for Mobile Robot, Homography, Image Representation, Pose Graph Optimization, Loop Closure, Opposite View Correspondence Detection, Registration.}

\abstract{Significant recent advances have been made in Visual Place Recognition (VPR), feature correspondence and localization due to deep-learning-based methods. However, existing approaches tend to address,  partially or fully, only one of two key challenges: viewpoint change and perceptual aliasing. In this paper, we present novel research that simultaneously addresses both challenges by combining deep-learnt features with geometric transformations based on domain knowledge about navigation on a ground-plane, without specialized hardware (e.g. downwards facing cameras, etc.). In particular, our integration of VPR with SLAM by leveraging the robustness of deep-learnt features and our homography-based extreme viewpoint invariance significantly boosts the performance of VPR, feature correspondence and pose graph sub-modules of the SLAM pipeline. We demonstrate a localization system capable of state-of-the-art performance despite perceptual aliasing and extreme 180-degree-rotated viewpoint change in a range of real-world and simulated experiments. Our system is  able to achieve early loop closures that prevent significant drifts in SLAM trajectories.}

\onecolumn \maketitle \normalsize \setcounter{footnote}{0} \vfill

\section{\uppercase{Introduction}}
\label{sec:introduction}
\noindent Visual Place Recognition (VPR) and local feature matching are an integral part of a visual SLAM system for correcting the drift in robot's trajectory via loop closures. However, multiple complicating factors make this process challenging such as variations in lighting and viewpoint along with the need to deal with dynamic objects.
% This has led to the researchers developing viewpoint-invariant~\cite{garg2018don} and change invariant methods~\cite{garg2018lost} for VPR. 

% In addition to all of these factors, 
Typically, indoor structures (e.g. walls, ceilings) tend to be feature-deficient. They often exhibit strong self-similarity, leading to perceptual aliasing. Ergo, VPR becomes further challenging when a place is revisited from a very different viewpoint eg. an opposing viewpoint ($180^{\circ}$ viewpoint shift). The latter is a situation commonly encountered when tackling VPR for indoor-based scenarios in warehouses, office buildings and their corridors.

Due to the above mentioned challenges, existing state-of-the-art place representation methods struggle to perform well. In particular, deep learning-enabled viewpoint-invariant global image representations~\cite{arandjelovic2016netvlad,garg2018lost} are unable to deal with perceptual aliasing due to repetitive indoor structures. Whereas, viewpoint-presumed image representations~\cite{dalal2005histograms} that retain spatial layout of the image fail due to $180^{\circ}$ viewpoint shift, as also demonstrated in~\cite{garg2018don}. Therefore, a robust place representation leveraging discriminative regions of an image is much needed to deal with this problem. 

% \Ravi{This line either needs to be removed or moved elsewhere. It proposes the final solution abruptly and too early into the narrative. ---
%  Such a representation can be achieved by combining deep-learnt features with geometric transformations based on reasonable domain assumptions about navigation on a ground-plane, thus leading to state-of-the-art localization performance.
%  }

 While seemingly aliased, in practice, floor patterns contain discriminative features. Blemishes, scratches on the floor surface and natural variations in floor/ground surfaces yield features which can be detected easily across conditional variations~\cite{zhang2019high}. In turn, these enable reliable localization~\cite{kelly2007field,nourani2009practical}. However, most of the existing solutions based on floor patches require specialised hardware (e.g. downward-facing cameras~\cite{nourani2009practical,mount2017image}, additional light sources~\cite{kelly2007field}).
 
%  \Ravi{This seems a strong claim. Can we back this up reliably ? --  Importantly, these methods have not been demonstrated to work under strong viewpoint variations.} 

We propose an indoor VPR approach which addresses the concerns highlighted previously. With our proposed pipeline, we make the following contributions:

\begin{itemize}
    \item A novel pipeline that combines projective geometry and deep-learnt features to focus specifically on floor areas and consistently improves the following pipeline modules: VPR (able to deal with opposing viewpoints), feature correspondences; leading to improved inputs for subsequent SLAM pose-graph optimization.
    
    \item Extensive comparisons across various deep architectures showing that VPR and feature correspondence modules suffer significantly when used on raw images while achieving a significant boost in performance when used on rotationally-aligned floor areas. The improvement is consistent over various real and simulated floor types as shown in section~\ref{sec:results}.
    
    % \item Extensive comparison over a variety of deep architectures, showcasing the superiority of NetVLAD~\cite{arandjelovic2016netvlad} to recognize places and D2-Net to extract dense feature correspondences \textit{only when using} rotationally-aligned opposite floor views. Both place retrieval and local feature correspondences across opposite views are critical for pose estimation for the SLAM back-end.

    \item The paper unveils Early Bird SLAM that integrates the above VPR and feature-correspondence pipeline in a back-end pose graph optimizer, demonstrating substantial decrease in Absolute Trajectory Error (ATE) as compared to the state-of-the-art SLAM frameworks such as~\cite{Rtabmap}.
    %which is unable to detect any loops from opposing viewpoints.
    
    %\item A comprehensive evaluation of the whole pipeline using a multitude of deep architectures and a variety of real-world and simulated datasets with differing floor textures, smoothness such as stone, marble, concrete, and wooden that verify the repeatability of the proposed pipeline.
    
\end{itemize}
\begin{figure*}
\centering
        \includegraphics[scale=0.245]{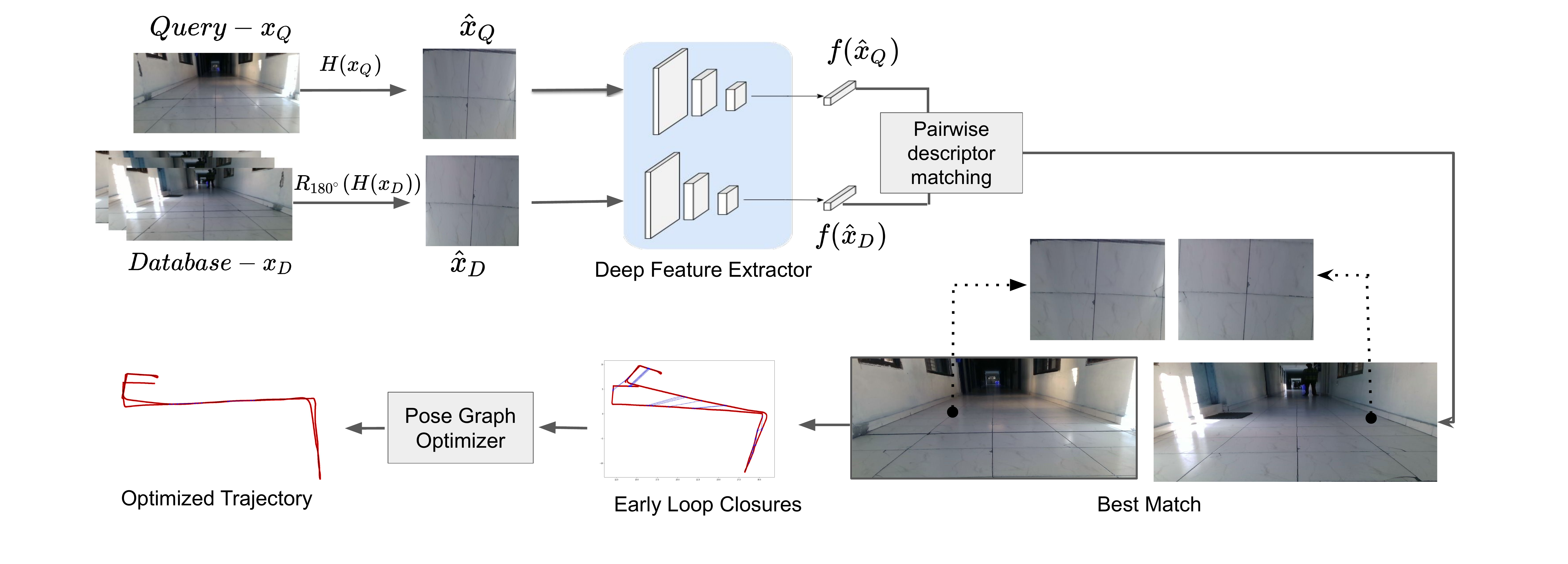}  
    \caption{Our proposed pipeline for Visual Place Recognition. Images are first converted to a top view by homography. This transformed view is fed into a deep feature extractor to obtain the output descriptor which forms our place representation. Then, a cosine-distance based cost matrix is constructed to find matches. A matched pair is then fed into a feature correspondence extractor to find correspondences. This is subsequently fed into a pose graph optimizer to obtain optimized trajectory.}
    \label{fig:pipelinez}
\end{figure*}
\section{\uppercase{Related Work}}

\subsection{Descriptor based recognition}
Amongst earlier methods, the most popular were appearance-based place descriptors such as Bag of Visual Words (BoVW)~\cite{sivic2003video,csurka2004visual} and Vector of Locally Aggregated Descriptors (VLAD)~\cite{jegou2010aggregating} where a visual vocabulary is constructed using local features like SURF~\cite{bay2008speeded} and SIFT~\cite{lowe1999object}. These have been used in FAB-MAP~\cite{cummins2008fab} and ORB-SLAM~\cite{mur2015orb} to good effect. 

Whole-image descriptors like Gist~\cite{oliva2006building} and HoG~\cite{dalal2005histograms}  presume the scene viewpoint to remain similar across subsequent visits of the environment, enabling VPR under extreme appearance variations as demonstrated in SeqSLAM~\cite{milford2012seqslam}.

\subsection{Robustifying VPR}
% One of the main challenges is robustifying VPR. Variations of lighting, viewpoint and  sensor parameters make VPR extremely challenging. This has led to researchers exploring viewpoint-invariant~\cite{garg2018don}, appearance-invariant methods~\cite{garg2018lost}.
Many solutions have been proposed to robustify VPR to viewpoint variation.

CNNs with their partial viewpoint invariance have been shown to robustify VPR (~\cite{sunderhauf2015place}). They allow for end-to-end training where one can in addition to using off-the-shelf  networks~\cite{arandjelovic2016netvlad}, train the later layers to obtain task/dataset specific results~\cite{radenovic2018fine}.  In~\cite{dlvpr}, pyramid pooling was shown to improve viewpoint robustness. 

\paragraph{Opposing Viewpoints}
Most of the existing literature that addresses viewpoint-invariance for VPR assumes a large amount of visual overlap.

LoST~\cite{garg2018lost} used dense semantic information to represent places and extract keypoints from within the CNN to enable high-performance VPR. This was improved upon in~\cite{garg2019look} using a topo-metric representation of places. In the vein of utilizing higher-order semantics, X-view~\cite{gawel2018x} uses dense semantic segmentation and graph-based random walks to perform VPR. %\cite{kim2017satellite} uses satellite imagery to perform localization by matching aerial and ground view data.

\paragraph{Saliency of floor features}  
Floor features have been shown to be salient enough to aid in VPR. In~\cite{zhang2019high}, features are extracted from floor surfaces to perform global localization. The imperfections in the tiles provide enough features that keypoints can be extracted.In~\cite{mount2017image} the authors have explored surface based localization methods for match verification using ground-based imagery. \cite{kelly2007field} proposed to use floor patches to perform local region matching in order to develop an infrastructure-free localization system. In~\cite{nourani2009practical}, authors developed a visual odometry system based on floor patches.

\paragraph{Keypoint Correspondences}
 Classical approaches like SIFT~\cite{SIFT} and SURF~\cite{bay2008speeded} tackle the problem of calculating the pixel level correspondences between images in a two-way approach by first detecting the keypoints and then describing a local region around the keypoint. Recent learning-based approaches like SuperPoint~\cite{superpoint} and D2Net~\cite{Dusmanu2019CVPR} combine detection and description by simultaneously optimizing for both the tasks. However, none of the approaches work well when deployed in perceptually-aliased and low-textured indoor settings, particularly when viewing a scene from an opposite direction. We show that the existing deep-learnt feature correspondence methods can lead to better matching by using certain regions of the image like floor and exploiting geometric priors between images in the forward and reverse trajectory.

\section{Methodology}

\begin{figure*}
\centering
        \includegraphics[scale=0.28]{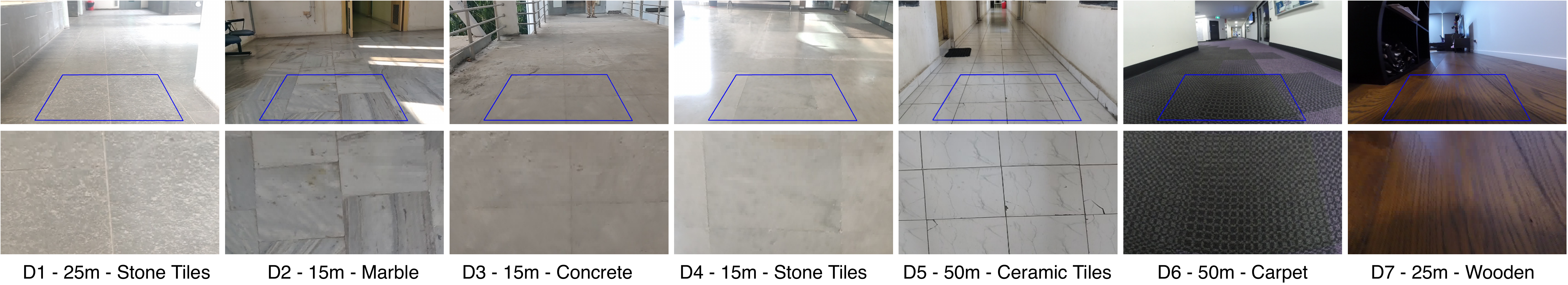}  
    \caption{Example images from our indoor dataset. We demonstrate the usage of our indoor VPR pipeline on various floor types.(First row) Raw images. (Second row) Homography transformed images of the respective raw images in the first row. Below we also mention the dataset id along with its trajectory length in meters and floor type in the form of (ID-Length-Floor Type). We overcome challenges such as self repetition and lack of features by using homography transformed floor images which contain distinctive features that are helpful in performing VPR. }
    \label{fig:homography-extraction-example}
\end{figure*}

%Building on the general approach of framing VPR as an image retrieval problem, we propose a method for VPR that deals with opposing viewpoints by aggregating the expressive power of deep-learned features into floor homography.
\noindent Our proposed hierarchical pipeline consists of the following three stages: indoor visual place recognition for opposite viewpoints, feature correspondence extraction and pose graph optimization.

\subsection{Indoor visual place recognition for opposite viewpoints}
\label{ssec:deepivpr}

% a shortest path problem that incorporates temporal constraints. 
% The shortest path problem allows for the incorporation of temporal consistency which we have found gives a significant boost to recall.

% One of the unique features of our method is that 
While the method employs deep-learned features, it requires no training of the underlying feature extractor.
% Using the features extracted from a Resnet-50 network pre-trained on the Imagenet dataset~\cite{deng2009imagenet} without any additional fine-tuning gives us good recall. 

% As mentioned earlier, our method is developed for, but not restricted to, VPR from opposing viewpoints. This is framed as a shortest path problem between the first and last frames under consideration.  
%As shown in~\cref{fig:pipelinez}, the proposed approach has three main components: (i) \emph{Homography Extraction} where floor patches from a forward facing camera are extracted via homography, (ii) \emph{Feature Extraction} where the homography-transformed images are passed through a pre-trained CNN to extract features which are dimension-reduced using PCA, and (iii) \textit{Loop Closure} where we formulate the data association constraints for pose graph optimization.
% (iv) \emph{Shortest-Path Computation}~\ref{ssec:shortest-path} we find a matching between query and reference images by solving a shortest path problem on a distance matrix constructed from the descriptors obtained in ~\ref{ssec:feature-extraction}. 

% The aggregating of temporal information is done in solving the shortest path problem resulting in temporal consistency.
                    
Our approach to indoor VPR utilises the fact that floor patches contain useful features in the form of cracks, designs, dirt/stains. The floor-based features act as a unique signature for specific places within an indoor region.
To extract the floor-region of the images taken, we fit a planar homography $H$ to image points via a RANSAC + 4 point algorithm~\cite{hartley2003multiple}.
% We use homography based extraction in two different scenarios. In the first scenario, the backward path is traced back exactly in the line of the forward trajectory, implying no lateral shift between the forward and backward trajectories. In the second scenario, we consider the case with some lateral offset between the forward and the backward trajectory. 
%\subsubsection{Homography estimation \textit{without} lateral shift}
%\label{sssec:nolshomography}
We use a fixed homography matrix across all the datasets. In the original image, we pick four points along the floor region which are then transformed into a floor image. 

%\subsubsection{Homography estimation \textit{with} lateral shift}
%\label{sssec:lshomography}
% In this case, for every dataset, we obtain a separate homography matrix for query traverse $\mathtt{Q}$ and database traverse $\mathtt{D}$. We follow a sliding window-based approach to search for homography matrix. We select a window near the center of the floor in the original RGB image and slide the window by 100 pixels each time along the horizontal axis. For each such window, there exists a homography which gives us a floor patch. For every dataset, we experiment with all the pairs of homographies corresponding to $\mathtt{Q}$ and $\mathtt{D}$. 
% We choose the pair of homographies which give us the highest recall value. 

Let $H$ be the homography matrix, ${x}$ be a homogenized coordinate of an input image then the transformed image co-ordinate, $\hat{x}$ is obtained via ~\cref{eq:homography}.
\begin{align}
    \label{eq:homography}
    \hat{x} &= H(x)
\end{align} 

Figure~\ref{fig:homography-extraction-example} shows example images from our benchmark datasets with both the raw images and their corresponding floor patches so obtained.

In our pipeline, we pass the floor patch images obtained into a deep feature extractor and the output descriptor forms our place representation.

\begin{align}
    d^{Q}_{i} &= f(\hat{x}^{Q}_{i})
    \label{eq:feature-extraction-pca}
\end{align} 
where $f()$ corresponds to the process of obtaining features from a deep feature extractor and $d^{Q}_{i}$ is the resultant descriptor obtained for image $i$. Cosine distance-based descriptor matching is done to obtain matches between $Q$ and $D$. We apply the homography operation on the reference and then perform a 180$^\circ$ rotation of the transformed images to improve matching across differing viewpoints of the same place. Although the rotation/flipping operation is not necessary for some of the deep feature extractors as they are inherently viewpoint-invariant, we show that performance can be boosted for such descriptors whereas other viewpoint-presumed deep feature description techniques become only useful post image rotation.

\subsection{Feature correspondence}
% Calculating the pixel level correspondences between images is a well-studied problem in computer vision for tasks like tracking and localization. Classical approaches like SIFT~\cite{SIFT} and SURF~\cite{bay2008speeded} tackle this problem in a two-way approach by first detecting the keypoints and then describing a local region around the keypoint. Recent learning-based approaches like SuperPoint~\cite{superpoint} and D2Net~\cite{Dusmanu2019CVPR} combine detection and description by simultaneously optimizing for both the tasks. 
% Still both classical and learning-based approaches fail in establishing correspondences between opposite-viewpoint images in low-texture regions in indoor settings, Figure~\ref{fig:correspondences} first and third row. SIFT is rotation invariant to very slight viewpoint changes. SuperPoint has been trained by generating random homography of a image while D2Net has been trained on outdoor SFM dataset by utilizing overlap between point clouds. Thus lack of opposite viewpoint images during training generates very imperfect matches when deploy in our use case.
%\SG{This para needs to make the point clear about the following reasons for correspondence failure: limited visual overlap due to opposing viewpoints, high perceptual aliasing due to repetitive structures and lack of rotation-invariance due to design choices of hand-crafting (SIFT) and training regimes (SP,D2N).} \ud{Done}

%\Madhav{Request Sourav to make the above para clear by adding what he has mentioned}

By utilizing the previously proposed concept of applying geometric transformations on image to extract textured floor regions enables us to generate very precise pixel level correspondences Figure~\ref{fig:correspondences} (1c and 2c). These precise correspondences are also very essential to calculate near ground truth transformation and subsequent loop closure in pose graph SLAM Figure~\ref{fig:Loop closure slam} on real dataset and Figure~\ref{fig:Loop closure slam-2} on synthetic dataset.
%\SG{The reference to first and third row in this and previous para is not a clear one as that figure has multiple lines/blocks.} \ud{Done}

% \SG{Figure~\ref{fig:correspondences} is an important one. If time allows, we can also show the floor correspondences back projected in the full image.}

\begin{figure*}
\centering
  \includegraphics[scale=0.19]{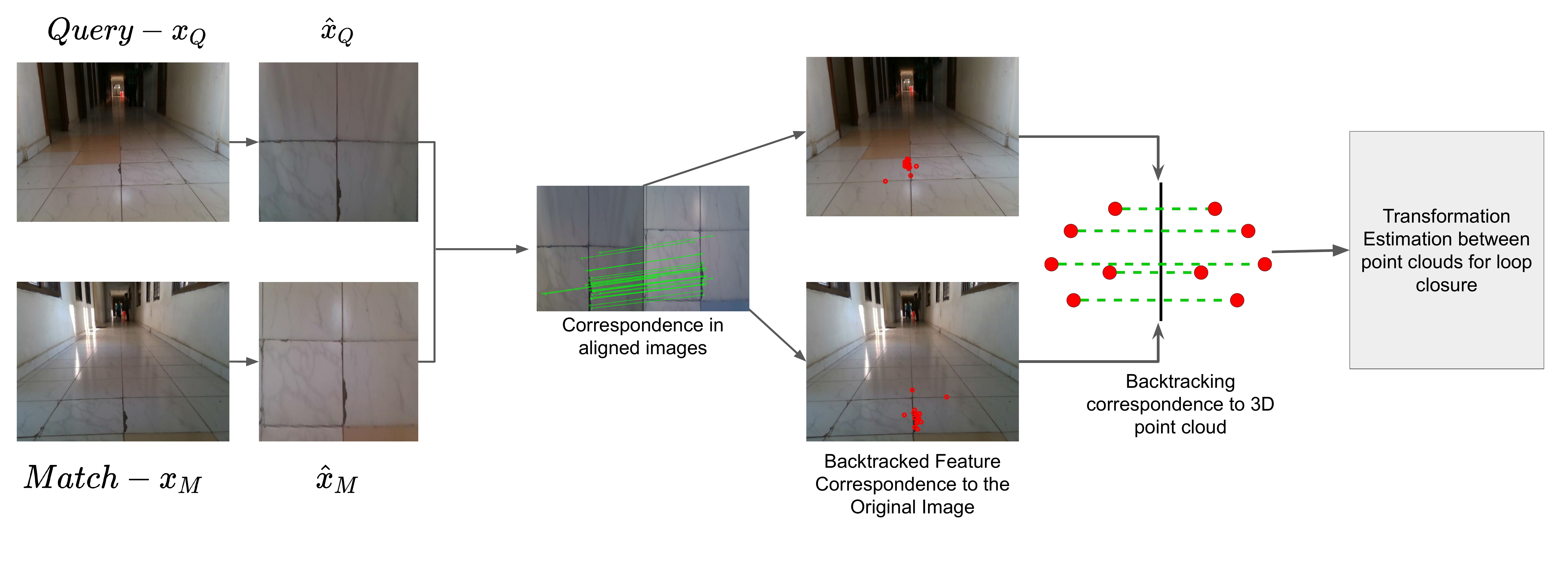}
  \caption{Pipeline depicting the process of transformation estimation from proposed opposite pairs by VPR.}
  \label{fig:Loop closure pipeline}
\end{figure*}

Let ${x}_{Q}$ be the query image and ${x}_{M}$ be the matched image from the opposite trajectory obtained via the VPR pipeline. $\hat{x}_{Q}$ is the transformed image obtained by applying homography and $\hat{x}_{M}$ is the transformed image obtained by applying homography and $\pi$-rotation, Figure~\ref{fig:Loop closure pipeline}. Local feature extractor $g(.)$, in our case D2Net is used to obtain correspondences $\hat{q}^{2D}$ and $\hat{m}^{2D}$ on transformed images. Correspondences on the original image $q^{2D}$ and $m^{2D}$ are obtained by inverse $\pi$-rotation and inverse homography. 

\begin{align}
\small
    \hat{x}_{Q} &= H(x_{Q}) \\
    \hat{x}_{M} &= R_{\pi}(H(x_{M})) \\
    \hat{q}^{2D}, \hat{m}^{2D} &= g(\hat{x}_{Q}, \hat{x}_{M}) \\
    q^{2D} &= H^{-1}(\hat{q}^{2D}) \\
    m^{2D} &= H^{-1}(R_{\pi}^{-1}(\hat{m}^{2D}))
\end{align}
 \subsection{Pose graph optimization}

The proposed VPR pipeline has direct applicability in loop closure or data association problem in visual SLAM. Formally, we are interested in finding the optimal configuration $X^{*}$ of robot poses $x_{i}$ based on odometry constraints $u_{i}$ and loop closing constraints $c_{qm}$. Here, odometry constraints $u_{i}$ are used to build the motion model whereas loop closure constraints $c_{qm}$ provide information to correct the error accumulated due to sensors' noise. 

Let $S$ be a set of image pairs proposed by VPR such that, $S = \{(q, m) | I_{q} \in Q, I_{m} \in R\}$, then optimal poses $\mathtt{X^{*}}$ are given by:

% \begin{align}
    
%     X^{*} = \underset{X}{argmax}\medspace P(X|U, C)
%     = \underset{X}{argmax}\medspace \underset{\textit{Odometry Constraints}}{\underbrace{\prod_{i}P(x_{i+1} | x_{i}, u_{i})}} \notag\\ 
%     \times \underset{\textit{Loop Closure by VPR}}{\underbrace{\prod_{(q, m) \in S}P(x_{m} | x_{q}, c_{qm})}}
    
% \end{align}

\begin{align}
\small
    X^{*} = \underset{X}{argmax}\medspace P(X|U,C)
    = \underset{X}{argmax}\medspace \underset{\textit{Odometry Constraints}}{\underbrace{\prod_{i}P(x_{i+1} | x_{i}, u_{i})}} \notag\\ \times \underset{\textit{Loop Closure by VPR}}{\underbrace{\prod_{(q, m) \in S}P(x_{m} | x_{q}, c_{qm})}}
\end{align}

% Obtain pixel-level correspondences between a pair of images is a well studied problem in the field of 3D computer vision, localization and tracking. Approaches like ~\cite{superpoint} and ~\cite{Dusmanu2019CVPR} make use of combining the process of detection and descriptions as they share the same underlying representation. We choose D2Net as it is invariant to scale changes as it uses an image pyramid scheme and performs well under large viewpoint changes. 

% Let $\mathtt{F_{AD}}$ be the aligned database image obtained by homography $\mathtt{H}$ and flipping $\mathtt{R_{180^{0}}}$ operations, that best matches with the query image $\mathtt{F_{Q}}$. Descriptor matching over D2-Net descriptor are done between $\mathtt{F_{AD}}$ and $\mathtt{F_{Q}}$ to obtain correspondences $\mathtt{(P^{2D}, Q^{2D})}$. The matched correspondences are tracked back to the original database image, $\mathtt{F_{D}}$ which is $180^{0}$ apart from the query $\mathtt{F_{Q}}$. The backtracking is achieved by inverse flipping and inverse homography operations. The complete pipeline to estimate transformation from image pairs is shown in \cref{fig:Loop closure pipeline}.

2D correspondences $q^{2D}$ and $m^{2D}$ obtained using D2Net in the previous subsection, are projected into $3D$ using the camera matrix $K$ and depth $\lambda$, and finally $3D$ points are registered using ICP. The ICP recovered transform  $c_{qm}$ between the two images form the loop closure constraint in the pose graph optimizer.

\begin{align}
    Q^{3D} &= \lambda K^{-1} q^{2D} \\
    M^{3D} &= \lambda K^{-1} m^{2D} \\
    c_{qm} &=    <R, T>  =   ICP(Q^{3D}, M^{3D})
\end{align}

% To obtain the transformation $\mathtt{u_{ij}}$, we are using the same homography $\mathtt{H}$ and flipping $\mathtt{R_{180^{0}}}$ operations as mentioned in VPR section to obtain floor patches $\mathtt{F_{i}}$ and $\mathtt{F_{j}}$. These floor matches are then matched using SURF~\cite{bay2008speeded} to obtain point correspondences $\mathtt{(P^{2D}, Q^{2D})}$. We found that these local features were useful for reliable keypoint correspondences for a known matching pair of images, however when used for ranking reference images across the whole dataset, VPR performance degrades due to perceptual aliasing. 

% The point correspondences so obtained are back projected into $\mathtt{3D}$ to obtain correspondences $\mathtt{(P^{3D}, Q^{3D})}$, which are then used by ICP algorithm to obtain precise rotation and translation estimates. Thus transformation $\mathtt{u_{ij}}$ is given by:

% \begin{equation}
%     F_{i}, F_{j}    =  R_{180^{0}}(H(I_{i}, I{j}))
% \end{equation}
% \begin{equation}
%     P^{2D}, Q^{2D}    =  SURF(F_{i}, F{j})
% \end{equation}
% \begin{equation}
%     u_{ij}    =    <R, T>  =   ICP(P^{3D}, Q^{3D})
% \end{equation}

% \begin{figure}[!h]\centering
%   \includegraphics[width=\linewidth]{pics/graph (2).png}
%   \caption{SIFT and correspondence}
%   \label{fig:SIFT}
% \end{figure}
To ensure that only high-precision loop closure constraints are used in pose graph optimization, we shortlist query-reference image pairs based on their cosine distance. We used only top 20 loop closure pairs with lowest cosine distance. Cauchy robust kernel in cost function is used to minimize the effect of false positive loop closures that might have crept in pose graph, using pose graph optimizer g2o~\cite{kummerle2011_g2o}. 

\section{Experimental Setup}
\label{sec:expsetup}

%\Madhav{needs some facelift/overhaul. You can organize into A) %Datasets B) Qualitative Results C) Quantitative Results D) Ablations %with lateral shift etc E) Loop Closure}

% \Madhav{The results section does not bring/mention in too much %duality like us vs NetVLAD. We just focus on the pipeline. Datasets %section should say the considerations made while considering the %corridors (texture, appearance, tile area etc), it needs to be more %detailed. Ablations should portray the role of each part of the %pipeline on ResNet. This is very critial. Qualitative results should %show how the tiles are different and argue why the dataset is %challenging. Quantitative should have something to say about why %ResNet and NetVLAD are doing better than others apart from %discussing the tables.
% The strength of the paper lies in how well you write the Results %Section. So all energies to this} 

\paragraph{Datasets} We have collected seven real-world indoor datasets in our experiments as shown in Figure~\ref{fig:homography-extraction-example}. Six in a university campus and one inside a home. The datasets comprise different types of floor types like marble, wooden, concrete and carpet. The datasets consist of sequences in range of 15 m to 50 m. Each sequence contains anywhere between 500-4000 images. Five were collected using a OnePlus 6 and two were collected
using GoPro Hero 3+. We have shown the performance of our VPR pipeline in a SLAM framework, Figure~\ref{fig:Loop closure slam}, on one of the dataset collected on the university campus with P3DX robot equipped with RealSense D435 and wheel odometer. Ablation studies of the effect of loop closures on the ATE of SLAM pose graph are done on three synthetic datasets, Figure~\ref{fig:Loop closure slam-2} (last column), where floor tiles are chosen from real world images and P3DX noise model have been incorporated in the simulator odometry data.

% The features of the farther floor patches lost after homography.  

%\SG{Need to specify your evaluation protocol, for example, define accuracy. If you mean %recall instead then use the word recall (everywhere) which is more common in VPR literature, %defined as TruePositives/TotalPositives.}\Satyajit{Updated}

\subsection{Evaluation and Comparisons}
We use Recall as an evaluation metric for visual place recognition, defined as the ratio of true positives and total number of positives. A match is said to be a true positive if it lies within a localization radius of $\frac{1}{15}$th of the total length of the traversal of its ground truth. We compare various deep feature extractors under different input settings.

For the feature correspondence results, performance of different types of input transformations is shown qualitatively and quantitatively in
 Figure~\ref{fig:correspondences} and Table~\ref{quant-corres} respectively. A correspondence match is considered to be an inlier if its reprojection error calculated using ground truth  transformation is below a threshold. Comparison among classical approach SURF~\cite{bay2008speeded} and learning based approaches SuperPoint~\cite{superpoint} and D2Net~\cite{Dusmanu2019CVPR} has been shown in Table~\ref{quant-corres}.

We have compared \textit{Early Bird} loop closures with state of the art SLAM method RTABMAP~\cite{Rtabmap} in terms of Absolute Trajectory Error(ATE) on both real world and synthetic datasets.

% For the feature correspondence results, we choose the top 20 correspondences calculated between the query and the match based on cosine distance and count the correct correspondences among these. We do an ablation among classical approaches like SURF~\cite{bay2008speeded} and learning based approaches like SuperPoint~\cite{superpoint} and D2Net~\cite{Dusmanu2019CVPR}.

% We measure the Average Trajectory Error as our metric for evaluating loop closure results. The ATE first aligns the ground truth trajectory with the proposed trajectory and then evaluates directly the absolute pose differences. We compare our pipeline with state-of-the-art SLAM system RTAB-Map~\cite{Rtabmap}.

%For NetVLAD and ResNet, we also consider alternatives like the use of raw image as input with or without left-to-right image flipping~\cite{garg2018lost}. Lastly, we also compare against image cropping as opposed to homography-based floor patch extraction to show the influence of perspective on VPR performance. \SG{Once Table 1 is split in two, update this description and mention that a detailed ablation study is provided in supplementary material using all transforms on all networks, if we decide to do that.}

\begin{table}[H]
\caption{Quantitative Analysis: First column shows the preprocessing applied to the input. Second column shows types of transformation applied on the reference images. $\pi$-Rot and Flip-LR indicate a 180$^\circ$ rotation and horizontal left-to-right flipping of the reference image respectively. Homo + $\pi$-Rot gives the best results in most cases. NetVLAD is used as the deep feature extractor. D1-D7 are the datasets as mentioned in Figure~\ref{fig:homography-extraction-example}}
    %\centering
    \ra{1.3}
    \resizebox{\columnwidth}{!}{
    \begin{tabular}{ll @{\hskip 2\tabcolsep}lllllll}
        \toprule
       Input & OP & D1 & D2 & D3 & D4 & D5 & D6 & D7\\
        \bottomrule
          Raw  & None & 24.1 & 22.1 & 24.8 & 19.9 & 9.7 & 28.5 & 22.2\\
            Raw & Flip-L-R & 28.8 & 26.8 & 29.1 & 26.6 & 12.5 & \textbf{29.2} & 20.6\\
            Homo  & None & 60.7 & 62.7 & 61.3 & 70.4 & 40.1 & 12.2 & 15.1\\
            Homo  & $\pi$-Rot & \textbf{69.3} &\textbf{ 70.1} & \textbf{71.8} & \textbf{73.3} & \textbf{44.7} & 16.3 & \textbf{28.8}\\
            \bottomrule
    \end{tabular}}
    
    \label{netvlad-quant}
\end{table}
\section{Results}
\label{sec:results}
\noindent We show results for each of the components of our pipeline, particularly highlighting the effect of geometric transformations for both VPR and feature correspondences which ultimately contribute in improving the trajectory error for the SLAM back end. First, we show results for VPR with ablations across many descriptors and geometric transformations followed by qualitative and quantitative results for feature correspondence extraction. Finally, we compare our Early Bird SLAM pipeline with the state-of-the-art SLAM system RTABMAP in terms of Absolute Trajectory Error (ATE) on real and synthetic datasets.

\subsection{Visual Place Recognition}
Table~\ref{netvlad-quant} and \ref{tab:desc_res} show the recall performance for VPR using seven different datasets. While Table~\ref{netvlad-quant} highlights the effect of geometric transformations on a given place descriptors, NetVLAD in this case, Table~\ref{tab:desc_res} compares different descriptor types for the best performing geometric transformation, that is, \textit{Homo + $\pi$-Rot}.  It can be observed in Table~\ref{netvlad-quant} that using the raw images (\textit{Raw}) as input leads to inferior results for most of the datasets even when using the state-of-the-art viewpoint-invariant representation NetVLAD. The best performance is achieved achieved through 180-degree rotation of floor patches (\textit{Homo + $\pi$-Rot}) as compared to when using only the homography transformed input (\textit{Homo}). We also compute descriptors using horizontally-flipped images (\text{Flip L-R}) as used in~\cite{garg2018don,garg2018lost} for dealing with opposing viewpoints in outdoor environments. It can be observed that such a transformation does not lead to consistent performance gains. We attribute this to the repetitive and featureless nature of indoor environments. For \textit{D7}, a small performance gap is observed between using \textit{Raw} and geometrically transformed images (\textit{Homo + $\pi$-Rot}); this is due to the reduced aliasing because of availability of unique visual landmarks when using raw images. In \textit{D6}, raw images (\textit{Raw}) perform better than floor (\textit{Homo + $\pi$-Rot}) images due to the lack of sufficient visual features on the carpet floor. This limitation could potentially be overcome by using a joint VLAD aggregation of the whole image and the transformed image, and remains a future work.
\begin{table}[H]
\caption{Quantitative Analysis: We compare the performance of various deep feature extractors where planar homography and 180$^\circ$ rotation (Homo+$\pi$-rot) is applied to the database images. D1-D7 are the datasets as mentioned in Figure~\ref{fig:homography-extraction-example}}
    \centering
    \ra{1.0}\resizebox{\columnwidth}{!}{
    \begin{tabular}{l@{\hskip 7\tabcolsep}lllllll}
        \toprule
        Models & D1 & D2 & D3 & D4 & D5 & D6 & D7\\
        \bottomrule
        NetVLAD     & 69.3 & \textbf{70.1} & \textbf{71.8} & \textbf{73.3} & \textbf{44.7} & 16.3 & \textbf{28.8}\\
        Resnet      & \textbf{72.8} & 66.5 & 60.6 & 63.3 & 38.0 & 16.9 & 13.2\\
        VGG-19      & 71.6 & 63.0 & 56.0 & 61.1 & 32.5 & \textbf{17.6} & 5.0\\
        Superpoint  & 16.5 & 27.6 & 15.0 & 23.0 & 35.7 & 1.6 & 3.2\\
        \bottomrule
    \end{tabular}}
    \label{tab:desc_res}
\end{table}
In Table~\ref{tab:desc_res} we compare the performance of different feature extractors under the best input setting (\textit{Homo + $\pi$-Rot}). NetVLAD performs the best in most cases with ResNet being the second best. While NetVLAD is a viewpoint-invariant representation, ResNet-based feature extraction assumes the viewpoint to be the same after geometric transformations. A viewpoint-invariant representation has more advantages which is reflected in Table~\ref{tab:desc_res}. Nevertheless, due to high perceptual aliasing, geometric transformations are \textit{still} required before descriptor computation in order to achieve the best performance, as demonstrated in Table~\ref{netvlad-quant}. 

\newcommand{\scaleOne}{0.1}
\newcommand{\scaleTwo}{0.12}
%\begin{figure}[!htbp]
%\centering
%  \includegraphics[width=\linewidth]{pics/quali.pdf}
%  \caption{\textit{Qualitative results on Dataset 1 (top), 2 and 4 (bottom):} In each column the raw image and its floor patch is displayed. The left column shows the query images from an opposing viewpoint to that of the reference traverse, the middle column shows correct matches from reference that were obtained using ResNet (Homo+$\pi$-Rot), and the last column shows false matches obtained from NetVLAD (Full Image with no transformation). Due to significant perceptual aliasing, viewpoint-invariant representation (last column) leads to incorrect matches whereas using the proposed pipeline based on floor patches, places can be recognized correctly (middle column).}
 % \label{fig:resnetvlad}
%\end{figure}

\subsection{Feature Correspondence}
\begin{figure}[!htbp]

  \includegraphics[width=\linewidth]{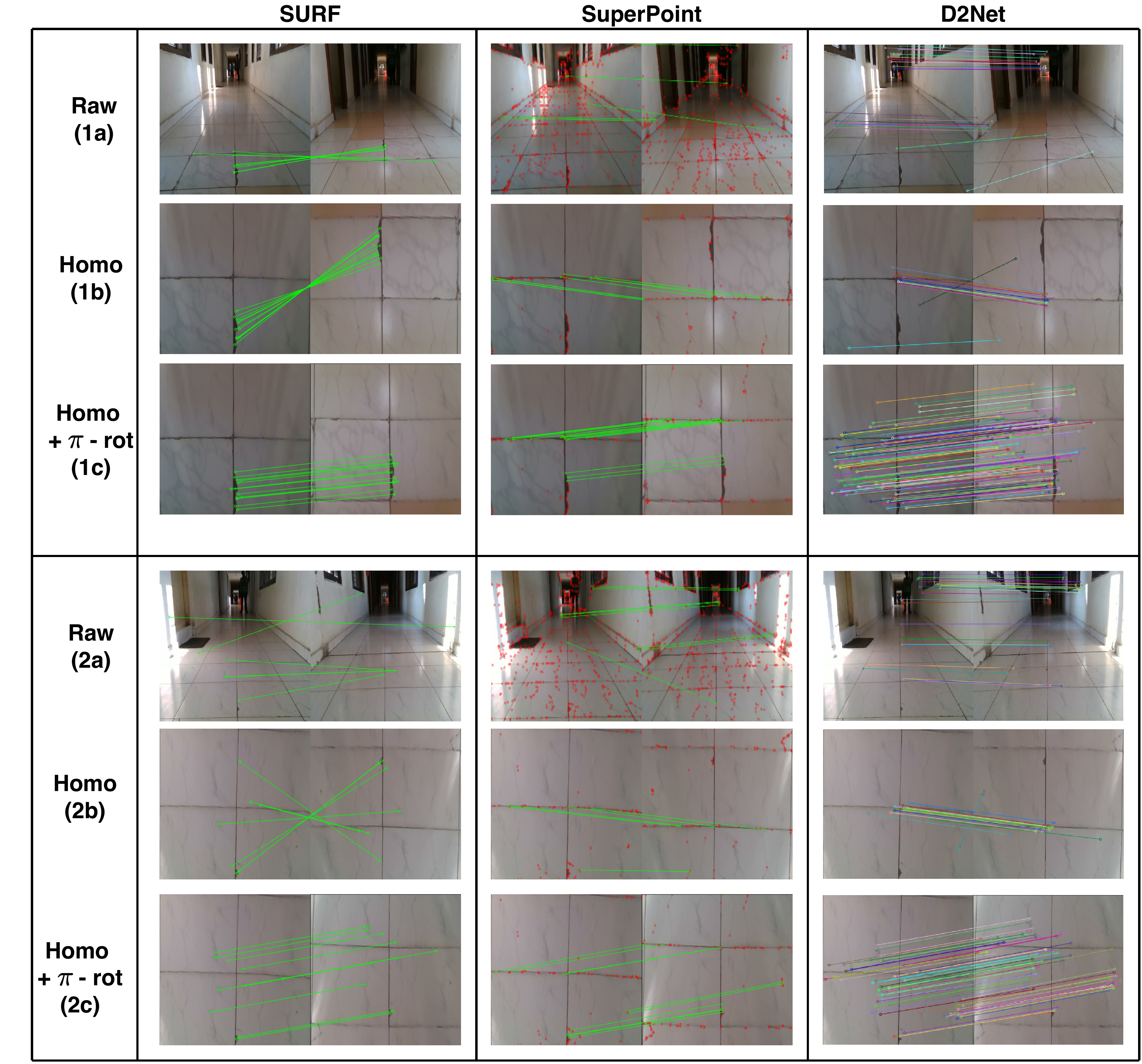}
  \caption{Comparison of correspondences obtained using raw image vs homo + $\pi$-rot. The best set of correspondences is obtained using D2Net with homo + $\pi$-rot operation.}% \ud{Done}
  \label{fig:correspondences}
\end{figure}

Estimating precise correspondences are crucial to calculate accurate transformations using ICP like registration methods, which in turn help us to achieve near ground truth pose estimates. Figure~\ref{fig:correspondences} (1a and 2a) show that using raw images to calculate feature correspondences cause both SURF and state-of-the-art learning methods SuperPoint and D2Net to fail. The number of correct correspondences increase with the use of geometric transformations focusing on textured floor regions. However, without image rotation, matching still remains poor as shown in Figure~\ref{fig:correspondences} (1b and 2b). The best results are obtained when transformed image pair is aligned with each other  by $180^{\circ}$ rotation as shown in Figure~\ref{fig:correspondences} (1c and 2c). Table~\ref{quant-corres} quantitatively shows the number of inliers as well as total correspondences, averaged over all the datasets, using different feature extractor methods. It can be observed that \textit{after} the geometric transformation, D2Net leads to a large number of initial as well as final correspondences. Thus, we used D2Net with homography and $\pi$-rotation as the final keypoint extractor approach for calculating transformations in subsequent tasks.

% Good correspondence matches are necessary to get accurate R and T between images which further aid in trajectory optimization. It can be observed from the figure results are inferior with raw images in all methods even in state-of-the-art methods like Superpoint as compared to homography transformed images due to repetitive and featureless nature in indoor environments. It can be observed from figure ~\ref{fig:correspondences}  with the homography and $\pi$-rot (\textit{Homo + $\pi$-rot}) we achieve significant performance gains in all the methods as compared with only homography transformed images (\textit{Homo}), In our case we are getting best results when combining our method with D2-Net as observed in figure~\ref{fig:correspondences}. In table ~\ref{quant-corres} we quantitatively compare the performance of classical and deep learnt approaches. When using Raw(\textit{Raw}) images as input we observe SURF to perform best with Superpoint and D2Net not being able to detect any inlier correspondences across datasets. D2Net is able to achieve the best performance when using Homography transformed images with (\textit{H}
%\label{quant-corres}

\begin{table}[ht]
 \caption{Number of Inliers / Total Correspondences averaged over all datasets.}
    \centering
    \ra{1.1}
    \scalebox{0.9}{
    \begin{tabular}{cccc}
        \toprule
         & SURF & Superpoint & D2Net\\
        \bottomrule
        Raw & 4.5 / 6.5  &   0 / 6.5    &  0 / 28.5\\
        Homo & 8.5 / 9  &   4 / 5.5   &  6.5 / 13\\
        Homo+$\pi$-Rot     & \textbf{11.5} / \textbf{11.5} &   \textbf{10} / \textbf{11.5}   &  \textbf{97.5} / \textbf{97.5} \\
        \bottomrule
    \end{tabular}}
   
    \label{quant-corres}
\end{table}
\begin{figure}[!htbp]

  \includegraphics[width=\linewidth]{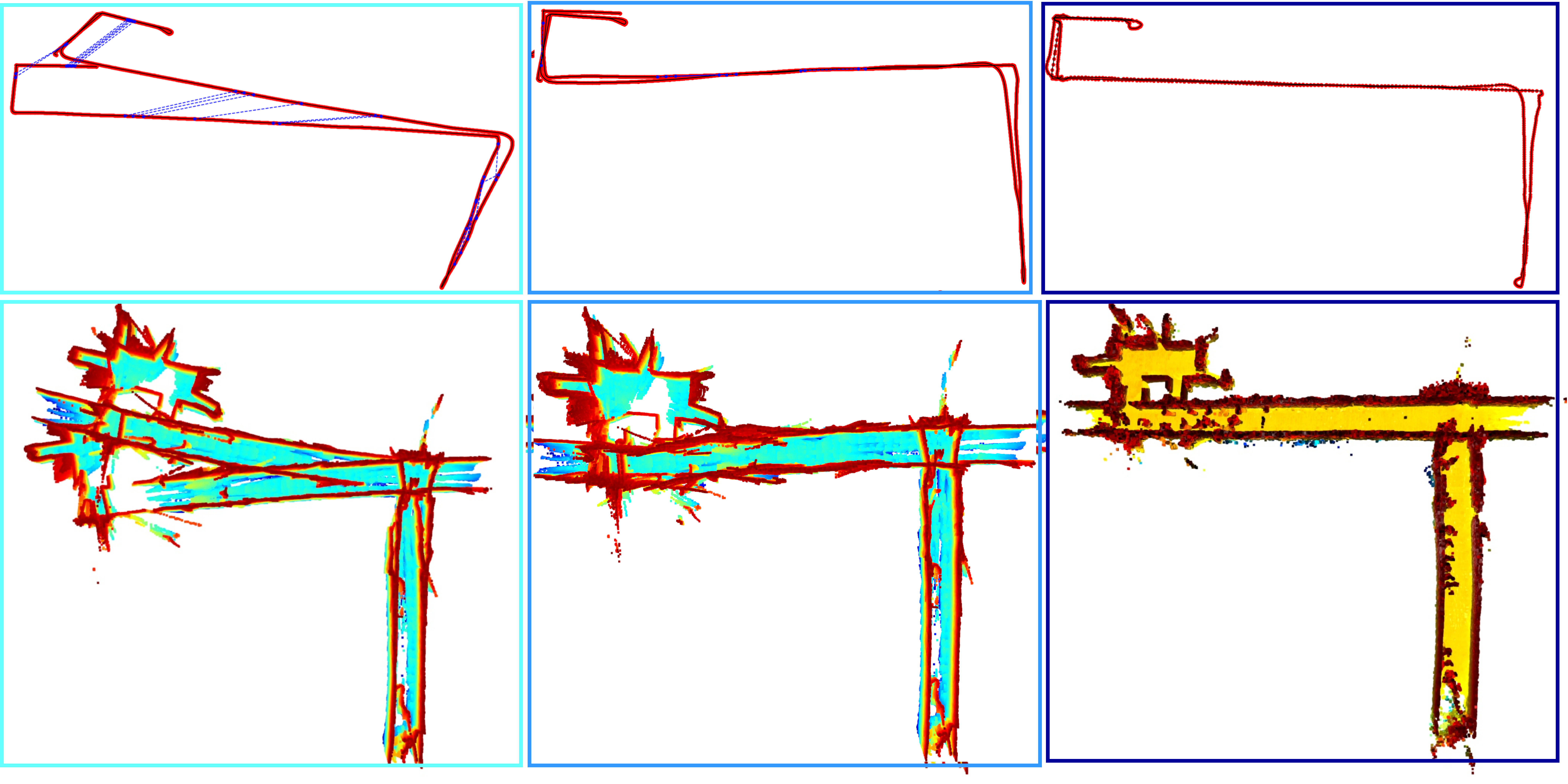}
  \caption{Rows represents pose graph and registered map.The first column corresponds to RTABMAP trajectory with robot revisiting the location from opposite viewpoints, blue dashed line represents early loop closures on pose graph, second and third column corresponds to optimized map based on VPR constraints and ground truth map respectively}
  \label{fig:Loop closure slam}
\end{figure}

\begin{figure}[!htbp]

  \includegraphics[width=\linewidth]{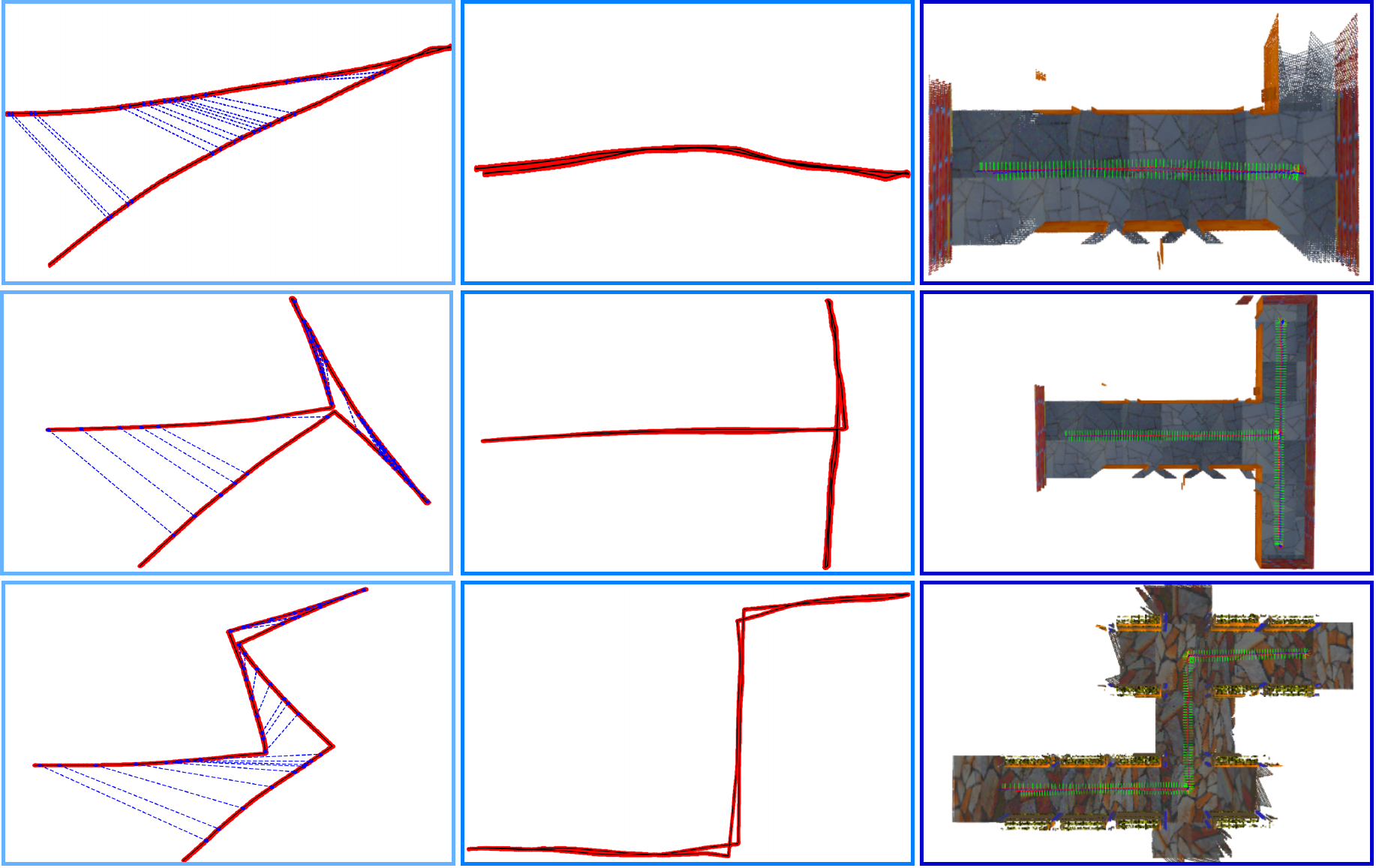}
  \caption{Each row represents pose graph optimization done on three gazebo environments corresponding to Table ~\ref{syntheticate}. First column corresponds to RTABMAP trajectory with robot revisiting the location from opposite viewpoints. Second and third column corresponds to optimized and ground truth trajectory.}
  \label{fig:Loop closure slam-2}
\end{figure}

\subsection{Loop Closures in Early Bird SLAM}

% We show the effectiveness of using floor features on three datasets. Each dataset consists of one example where the first two columns correspond to the original query image  and homography transformed query image. The next two columns show the original and homography transformed image of the match obtained from the reference database. Rotating the images of column four by  180$^\circ$ get spatially aligned with the images of column 2. It can be seen that both images are of the same region. The fifth column corresponds to the false match obtained from NetVLAD which takes original image as input without any transformation. For a clear visualization of the false matches obtained by NetVLAD we also show the homography transformed image of the matched images in column 6.
% \subsubsection{Benchmark against traditional SLAM system}
In practical scenarios in the context of long term autonomy, a robot can typically revisit its operating environment from a variety of different viewpoints. A more common scenario  particularly in corridors and aisles is that of an opposite viewpoint which is when our proposed system triggers loop closure. We demonstrate its efficacy by comparing with the state-of-the-art SLAM system RTAB-Map~\cite{Rtabmap}. As shown in the Table~\ref{syntheticate} (Dataset D5), we significantly reduce the Average Trajectory Error (ATE) by detecting “early" loop closures which is qualitatively shown in Figure~\ref{fig:Loop closure slam} while RTABMAP fails to do so, due to significant perceptual aliasing. As a consequence, the RTABMAP's trajectory Figure~\ref{fig:Loop closure slam}  (\textit{top left}) shows multiple corridors when actually there is only one. The \textit{(middle)} column  Figure~\ref{fig:Loop closure slam} shows trajectory and top view map due to loop detection and closure from opposite views using our VPR pipeline. These are much closer to ground truth trajectory and map shown in \textit{(last)} column of Figure~\ref{fig:Loop closure slam}. Ablation study has been done on measuring the usefulness of early loop closures with different robot's trajectory length and different tile patterns. We have performed our experiment in three different simulated environment settings as shown in Figure~\ref{fig:Loop closure slam-2}. In all the three cases, we have achieved near ground truth poses, while state of the art SLAM method RTABMAP fails to detect \textit{any} loop closure while returning to the same place from opposite viewpoints. These results are quantitatively represented in terms of ATE in Table~\ref{syntheticate}, showing that with increase in the length and complexity of the trajectory the effect of \textit{early} loop closures becomes even more pronounced.

\begin{table}[h]
\caption{Average Trajectory Error on a university dataset and three gazebo datasets for RTABMAP loop closures vs Early Bird loop closures.}
\begin{center}
% \scalebox{0.8}{
\begin{tabular}{lcc}

\toprule

Datasets & RTABMAP & Early Bird \\
\midrule
D5 & 9.239 & \textbf{5.69}\\
S1 & 0.94 &  \textbf{0.26}\\
S2 & 2.86 &  \textbf{0.38}\\
S3 & 2.85 &  \textbf{0.36}\\
% Ours & 6.035\\

% RTAB-Map & 9.239\\

\bottomrule

\end{tabular} 
% }

\label{syntheticate}
\end{center}

\end{table}

\section*{Conclusion and Future Work} 
\noindent This paper proposes a novel pipeline that integrates Visual Place Recognition (VPR) with SLAM front and back ends specifically for loop detection and closure for opposite views. The paper showcases that rotationally-aligned deep floor descriptors provide for significant boost in all the submodules of the pipeline: VPR (loop detection), descriptor matching and pose graph optimization. 

The paper extensively compares several deep architectures for VPR and descriptor matching. Given the geometric transformations, NetVLAD as a global descriptor was found most suitable for opposite-view VPR while descriptor matching through D2Net found maximal number of matches vis a vis competing descriptor matching frameworks. Also, superior place recognition and descriptor matching across opposite views resulted in a similar performance gain in back-end pose graph optimization. Specifically, we showed \textit{early} loop closures that prevented significant drifts in SLAM trajectories as a consequence of the proposed pipeline along with the proper choice of deep architectures that exploit the various sub-modules of the pipeline to its maximum efficacy. 

% The future threads include extension to outdoor and warehouse like topologies, use of visual semantics or monocular depth-based ground plane extraction. 
% and learning an attention mechanism to deal with the simultaneous effect of viewpoint variations and perceptual aliasing.

\bibliographystyle{apalike}
{\small
\bibliography{example, vpr}}

\end{document}